\documentclass[conference]{IEEEtran}
\IEEEoverridecommandlockouts
\usepackage{cite}
\usepackage{amsmath,amssymb,amsfonts}
\usepackage{algorithmic}
\usepackage{graphicx}
\usepackage{textcomp}
\usepackage{xcolor}
\def\BibTeX{{\rm B\kern-.05em{\sc i\kern-.025em b}\kern-.08em
		T\kern-.1667em\lower.7ex\hbox{E}\kern-.125emX}}
\usepackage{booktabs} % For formal tables
\usepackage{color,soul}
\usepackage{amssymb}
\usepackage{amsmath}

\begin{document}
\title{Focusing on the Big Picture: Insights into a Systems Approach to Deep Learning for Satellite Imagery}

\author{\IEEEauthorblockN{Ritwik Gupta, Carson D. Sestili\IEEEauthorrefmark{1}\thanks{\IEEEauthorrefmark{1}C. Sestili is not affiliated with the Software Engineering Institute at the time of publication and can be reached at csestili@alumni.cmu.edu},  Javier A. Vazquez-Trejo, Matthew E. Gaston}
	\IEEEauthorblockA{Software Engineering Institute\\
		Carnegie Mellon University\\
		Pittsburgh, USA\\
		\{rgupta, cdsestili, javazquez, megaston\}@sei.cmu.edu}
}

\maketitle

\begin{abstract}
Deep learning tasks are often complicated and require a variety of components working together efficiently to perform well. Due to the often large scale of these tasks, there is a necessity to iterate quickly in order to attempt a variety of methods and to find and fix bugs. While participating in IARPA's Functional Map of the World challenge, we identified challenges along the entire deep learning pipeline and found various solutions to these challenges. In this paper, we present the performance, engineering, and deep learning considerations with processing and modeling data, as well as underlying infrastructure considerations that support large-scale deep learning tasks. We also discuss insights and observations with regard to satellite imagery and deep learning for image classification.
\end{abstract}


\IEEEpeerreviewmaketitle

\maketitle

\section{Introduction}
Identifying the functional use of facilities and land in satellite images is a problem that experts from industry, academia, and government have explored and continue to explore in depth.
Due to the complex and heterogeneous nature of satellite imagery \cite{Lu2007}, even within categories, classification of land use within satellite imagery is a daunting task.
Furthermore, high-resolution datasets of satellite imagery are unwieldy to store, transfer, and manipulate, and performing complex learning tasks on such high-resolution data becomes complicated as systems run into storage, bandwidth, and memory issues.

During the IARPA Functional Map of the World challenge \cite{Christie2017}, we explored a difficult deep learning task in a competitive environment with computational restrictions.
This challenge necessitated an infrastructure that could support fast experimentation and agile, iterative development.
In this paper, we demonstrate and instrument limitations to infrastructure and deep learning methods and present our effective workarounds.
We describe 1) the underlying data and feature engineering, 2) the computational infrastructure supporting deep learning, and 3) deep learning approaches, their pros and cons, and attempted solutions to problems these approaches present.
\begin{figure}[h]
	\includegraphics[width=7in, height=1in, keepaspectratio]{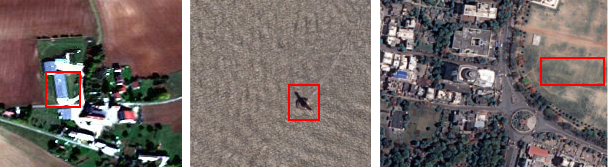}
	\caption{Example images from the dataset with bounding boxes manually superimposed: bounding box around a barn (left); bounding box around a lighthouse (center); example of a false detection (right).}
\end{figure}

\section{Data processing}
\label{DataProcessing}
Satellite imagery is highly heterogeneous, with variations in landscape, structures, cloud cover, and more. Proper data processing is extremely important to learn useful features from the input data.

\subsection{Bounding Boxes and Context}
\begin{figure}[h]
	\centering
	\includegraphics[width=2in, height=1.3in, keepaspectratio]{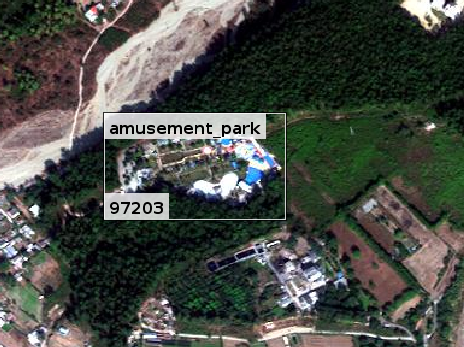}
	\caption{A tight bounding box around an amusement park.}
	\label{fig:BoundingBox}
\end{figure}

The metadata provided in the dataset for the IARPA Functional Map of the World (FMoW) challenge defines bounding boxes within the images to be classified. These bounding boxes are usually tight around the area of interest (Figure \ref{fig:BoundingBox}). It would be simple to present only the bounded portion of the image to a model. However, our hypothesis was that the landscape surrounding the bounded portion of the image gave important context that would improve the representation of the overall image. Therefore, we needed a way to expand the bounding box to look at the \textit{context}. To do this, we defined a method to create a context window and performed experiments to determine a reasonable context window size.

\begin{figure}
	\centering
	\includegraphics[height=1.3in, keepaspectratio]{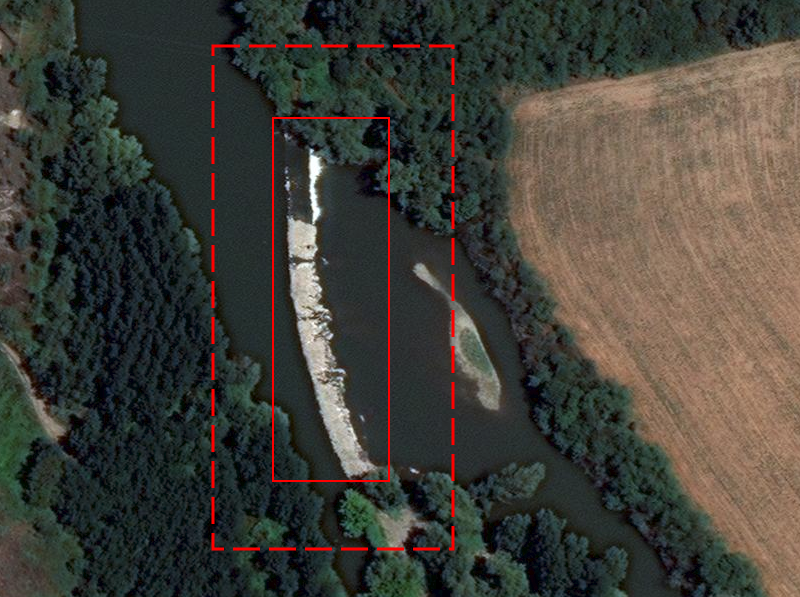}
	\caption{Example of a bounding box and the expanded context window around an instance of ``dam."}
	\label{fig:ContextWindow}
\end{figure}

\begin{figure}
	\centering
	\includegraphics[width=3in, height=2in, keepaspectratio]{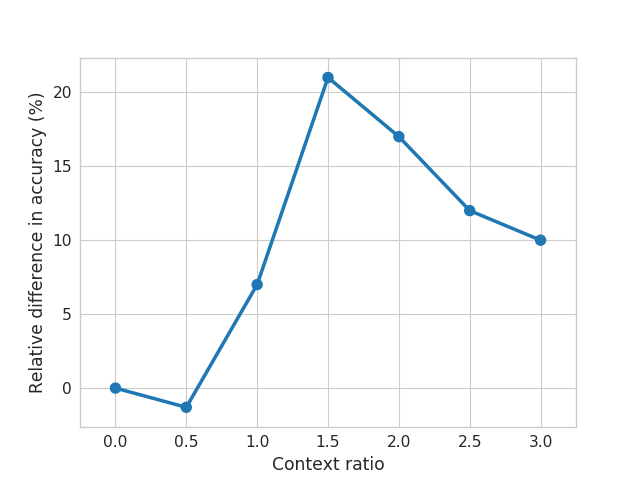}
	\caption{Relative difference in accuracy with varying context ratio.}
	\label{fig:ContextRatioDifference}
\end{figure}

We define the context window around the bounding box to be dependent on a \textit{context ratio}, $C$, to be
$$context\_window = \frac{C * AR}{2}$$ where $AR$ is the aspect ratio of the image. The bounding box would be expanded to cover an extra $context\_window * width$ pixels, and by the same factor for the height (see Figure \ref{fig:ContextWindow}). A context ratio of 0 would imply no extra context window. In order to decide on an optimal context ratio, we performed an experiment where we trained a simple convolutional neural network on the input data while varying the context ratio. We observed that a context ratio of {\raise.17ex\hbox{$\scriptstyle\sim$}}{1.5} resulted in the greatest performance (see Figure \ref{fig:ContextRatioDifference}).

\subsection{Highly Variant Bounding Box Resolutions}
Images within and between categories varied in resolution. While some images, such as certain instances of ``airport", were upwards of 4000x4000 px, other categories, such as ``zoo", had images lower than 200x200 px. Furthermore, images rarely had an aspect ratio of 1, causing aspect ratio to be a case to consider when trying to handle images of different resolutions. Convolutional neural networks have no requirement that images must be the same resolution. However, fully connected layers rely on fixed vector sizes to operate, which imposes that convolutional feature maps should be of the same dimensions. We explored multiple strategies to work around this issue.

\subsubsection{Bounding Box Rescaling}
A common method of handling datasets with images of varying resolutions is to rescale all images to the same size \cite{Krizhevsky}. In the case of \cite{Krizhevsky}, all images were downsampled to a fixed common size (256x256px). However, due to the extremely large range of resolutions in our dataset, downsampling all images to a small size would cause large images to be affected much more than already small ones. To find a middle ground, we took a sample of images and computed their mean and median resolutions (Table \ref{tab:ImageResolutions}). Since the mean and median differed by a large margin, we chose to rescale images to a size close to the median to ensure that a large number of images would have to undergo small transformations. Furthermore, the chosen size had an aspect ratio of 1 to ensure that the convolution math in neural networks would be simple. By skewing the aspect ratio of the images, we force the models to learn an inaccurate representation of the underlying data. While this is not an ideal trade-off, it was one that did not have a considerable impact. A more suitable solution could be crafted with further exploration of this idea.
\begin{table}
  \caption{Sample Image Resolution}
  \label{tab:ImageResolutions}
  \centering
  \begin{tabular}{cccc}
    \toprule
    \space&Width (px)&Height (px)&Aspect Ratio\\
    \midrule
    Mean& 367& 289&{\raise.17ex\hbox{$\scriptstyle\sim$}}1.27\\
    Median& 245& 196& 1.25\\
  \bottomrule
\end{tabular}
\end{table}

\subsubsection{Spatial Pyramid Pooling}
Spatial pyramid pooling (SPP)\cite{He2015} uses a bag-of-words approach to create fixed-length vectors that maintain spatial information. This approach lets us replace the last pooling layer in the network with an SPP layer, which allows images of any input size to be fed into the model.
There are significant real-world limitations to this strategy. In \cite{He2015}, only images of two possible resolutions are used (i.e., all images are resized to either 180x180 or 224x224, depending on the closest dimension). We attempted to use this approach without any resizing, leaving bounding boxes in their original resolutions. In Keras, our framework of choice for this problem, tensor allocation on the GPU is fixed and not garbage collected until later in the execution cycle. As a result, tensors are created for every image that has a unique resolution, which results in an extremely rapid GPU memory exhaustion. Our training process terminated within a few batches because no further tensors could be allocated.

A possible acceptable trade-off to using SPP in a dataset, with large variation in width and height would be to create a limited amount of possible image resolutions based on summary statistics and resize images to the closest ``bucket." This approach would have ensured that our GPU could maintain the allocated tensors in memory and not distort images by a large amount. Due to time limitations, we were unable to test this methodology.

Spatial pyramid pooling has been empirically shown to boost the representation capacity of the network, leading to higher accuracy\cite{He2015}. Our hypothesis is that the use of such a technique would lead to large gains in accuracy on satellite imagery, which is naturally well-fit to the problem SPP attempts to solve.

\subsection{Different Pixel Scales}
\begin{figure}[h]
	\centering
	\includegraphics[width=3in, height=2.25in]{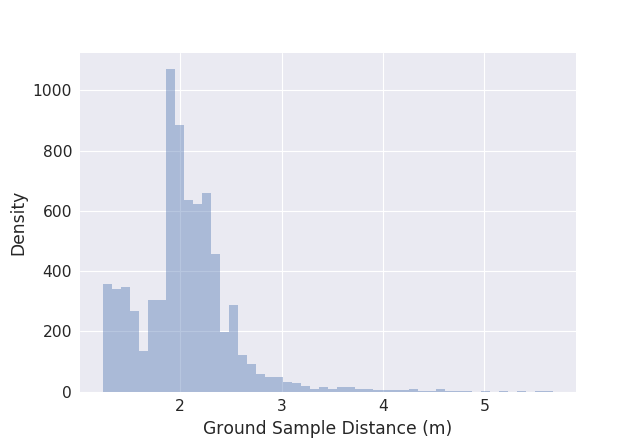}
	\caption{Distribution of ground sample distances in the dataset.}
        \label{fig:GroundSampleDistance}
\end{figure}

Every bounding box has an associated \textit{ground sample distance} (GSD), a field in the metadata that gives the side length, in meters, of the square on the Earth's surface that each pixel in the image represents. A wide range of GSD scales is present in the dataset (see Figure \ref{fig:GroundSampleDistance}).

Such different scales would fundamentally change the way the data is interpreted by a model. Therefore, we created a simple scaling mechanism to normalize all bounding boxes to the same scale so that every pixel represented a 1x1 meter square on the ground. Based on the distribution of ground sample distances in Figure \ref{fig:GroundSampleDistance}, a GSD of 1 meter is not an ideal rescaling target since the median GSD is at 2 meters. However, we chose a normalized GSD of 1 meter because it is a unit number and allowed us to use simpler calculations to perform the experiment.

Overall, normalizing the bounding boxes to a GSD scale of 1 meter provided a marginal boost in accuracy of {\raise.17ex\hbox{$\scriptstyle\sim$}}2\%. We strongly believe that normalizing to the median GSD would provide a larger boost in accuracy.

\subsection{Data Augmentation}
Data augmentation is a well explored practice to make neural networks more robust to various types of input transformations \cite{Krizhevsky, Hosseini}. We augmented both the metadata and the images to allow the models to learn a more representative function over the input space.

\subsubsection{Image augmentations}
We defined a set of basic transformations that would provide a large variety of alternate views on the data in the FMoW dataset (see Table \ref{tab:ImageAugmentations}).
\begin{itemize}
	\item{Rotations (15, 30, 45, 90, 180 degrees)}
	\item{Flips (East-West, North-South)}
	\item{Zooms (-1.5 - 1.5x)}
	\item{Channel-wise noise addition}
\end{itemize}
Performing the augmentations during the training process was a time-consuming operation. On large batch sizes, we saw slowdowns in training of up to 2.5x. In order to decrease training times to perform quicker iterations, all image augmentations were preprocessed and saved to disk ahead of time. This approach provided us with a rapid training pipeline as no expensive CPU processing was not necessary for every image in the dataset; such processing could add multiple hours onto training due to the amount of time it takes to perform these transformations. 

Train-time augmentation allows us to apply a random combination of transformations to each image whereas preprocessing augmentations provides a fixed set of transformations to feed through the model. A lower amount of combinations, in the case of preprocessing, results in lower generalization. This is a trade-off that must be examined carefully. Preprocessing all possible combinations is possible but requires an extremely large amount of storage, something we did not have. In our case, we were able to use the time saved in this part of the pipeline to discover methods that led to much larger increases in accuracy than train-time augmentation.
\begin{table}[h]
	\caption{Effects of Image Augmentations}
	\label{tab:ImageAugmentations}
	\centering
	\begin{tabular}{lc}
		\toprule
		Transformation & Model Accuracy\\
		\midrule
		None & 42\%\\
		Rotations & 53\%\\
		Rotations + Flips & 69\%\\
		Rotations + Flips + Zooms & 72\%\\
		Rotations + Flips + Zooms + Noise & 73\%\\
		\bottomrule
	\end{tabular}
\end{table}

\subsubsection{Image processing}
With the heavy number of data transformations that we perform on a large set of data, we had to choose a fast and performant library that could perform the necessary transformations. We considered a few Python libraries to use, and compared two in the end: PIL and OpenCV.

\begin{figure}
	\centering
	\includegraphics[width=3.5in, height=2in, keepaspectratio]{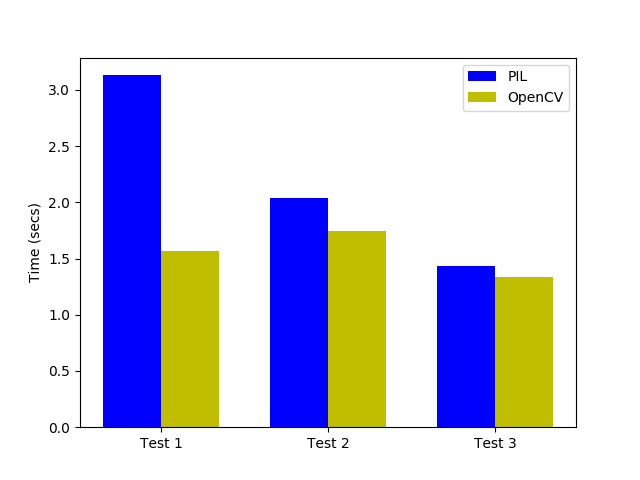}
	\caption{Comparison between PIL and OpenCV.}
	\label{fig:PILOpenCVCompare}
\end{figure}

We ran three main tests to compare performance between PIL and OpenCV. The tests were run on 300 images, averaged over 5 runs.
\begin{itemize}
	\item \textbf{Test 1:} Load an image, blur it, and flip
	\item \textbf{Test 2:} Load an image, rotate by 45 degrees
	\item \textbf{Test 3:} Load an image, rescale to a fixed size
\end{itemize}
The comparison results are shown in Figure \ref{fig:PILOpenCVCompare}. Overall, PIL performed slower than OpenCV, especially on flips, which account for a considerable portion of our augmentations. While PIL provides an easy-to-use API, our main focus was getting maximum performance from the data processing pipeline. When dealing with a large amount of data, the disparity between the frameworks quickly adds up to a noticeable difference in time. Since our processing pipeline was relatively simple and did not require complicated, detailed image manipulation, the slightly more verbose API of OpenCV was not an issue.

The overall trade-off is decided by the size of the dataset. On a smaller dataset, PIL and OpenCV do not noticeably differ on the amount of time spent processing images. However, on the FMoW dataset, the small differences in processing add up to over 1.5 hours. Since this is a sizable time sink to the goal of quick iterations, we chose OpenCV to implement our transformations.

\subsubsection{Metadata augmentations}
As an added challenge, IARPA introduced noise into their testing metadata. Therefore, in order to truly be robust to test-time input fluctuations, we had to ensure that the metadata was augmented to account for some uniform noise range around the true data. Therefore, we randomly created vectors of noise sampled from a uniform distribution with various ranges and added them to the normalized metadata vectors. Additionally, we added noise to the time and date in the metadata by sizable amounts to account for differences in time of day and year.

Since these operations are simple, we left these augmentations to be performed at training time. Unlike images, the metadata are simple vectors containing less than 50 elements. Because operations on these vectors are computationally simple, we did not incur any noticeable increase in training time.

Overall, metadata augmentation provided us with a {\raise.17ex\hbox{$\scriptstyle\sim$}}2-3\% increase in test accuracy.

\section{Deep Learning Discussion}
While working on this problem, we attempted to develop and use many different deep learning and machine learning techniques. In doing so, we observed some interesting behaviors and discovered some intriguing problems that are worth discussing.

\subsection{Pre-trained Models}
Initializing models with weights trained on large datasets such as ImageNet has been shown to be unreasonably effective \cite{Oquab2014, Marmanis2016}. It was previously believed that the massive size and diversity of ImageNet created a network that learned general features, but this hypothesis was shown to be flawed \cite{Huh2016}. In our experiments, ImageNet pre-training demonstrated dramatic improvements in accuracy. However, we hypothesized that pre-training on a dataset more similar to our actual satellite imagery dataset would result in better overall accuracy compared to ImageNet.

We used a simple VGGNet \cite{Simonyan2015} convolutional neural network with no pre-trained weights as a baseline model. The model was trained on the IARPA dataset, and the resultant accuracy was logged. We repeated the experiment with VGGNet pre-trained on ImageNet as well as DeepSat \cite{Basu2015} datasets. As expected, using pre-trained weights resulted in a massive gain in accuracy as opposed to not pre-training at all. Furthermore, we observed a {\raise.17ex\hbox{$\scriptstyle\sim$}}5\% gain in accuracy when using DeepSat to pre-train the model as opposed to ImageNet. Due to the limited amount of trials and detailed methodology, we are cautious about making claims regarding the efficacy of this solution, especially with regard to deeper, complicated models. Given more time, we would like to further explore the use of DeepSat and compare the differences in accuracy, precision, and recall with models pre-trained on other datasets. Related work in transfer learning gives credence to this idea \cite{Pan2010}.

\subsection{False Detections}
A sizable part of this challenge was accurately classifying false detections. Incorrectly handling false detections -- the 4th most populated class in the dataset -- significantly affected overall precision and recall. Samples of false detections were not provided in the training data, but they were present in validation and testing data. In literature, this problem is known as selective classification or a reject option \cite{Geifman2017, Nadeem2010}. We had multiple options for handling this problem.
\subsubsection{Data mixing}
A simple solution was to take a very limited subset of false detections provided in the validation dataset and use them to train the model. However, due to the low availability of data, this solution would have led to limited generalization and higher confusion across the entire network. \cite{Christie2017} used this approach to achieve an acceptable gain in accuracy. As such, data mixing should only be used when larger amounts of data are available to provide a comprehensive look at false detections.
\subsubsection{Random cropping}
\cite{Christie2017} do not provide a detailed explanation of how false detections were labeled in the dataset. Without a proper methodology to create false detections, an option would be to randomly crop images containing other classes in regions that are not enclosed by existing bounding boxes. This option would provide us with a false detection dataset as big as we required and would give us greater generalizability across the network. However, a significant shortcoming of this method is that a random crop may encompass an instance of another valid category. As a result, we may create ``false detections" that are mislabeled.

After attempting these possible solutions individually as well as a combination of the two, we compared the precision and recall between not handling the false detection problem and our attempted solutions. Overall, the attempted solutions resulted in a greater confusion across all categories as opposed to not attempting a solution at all. We speculate that these techniques would show greater positive effect on precision and recall with further development.

\subsection{Neural Network Architecture Design}

Our neural network models used multiple CNNs in parallel as components, including both a network pre-trained on ImageNet and an initially untrained network. These two networks function as simultaneous feature extractors, and we combine their features by concatenating their outputs and feeding the result to a network of several dense layers. The motivation behind this choice is that while the network that was pre-trained on ImageNet can already extract features from natural images, the initially untrained network can learn to extract features from the problem-specific dataset of satellite images.

To optimize the satellite image feature extractor sub-network, we experimented with various convolutional neural network architectures, including VGGNet \cite{Simonyan2015}, ResNet \cite{he2015b}, and DenseNet \cite{huang2016}. We found that the best performance occurred with the DenseNet161 architecture, whose performance we compare against other architectures in Table \ref{tab:arch}. It is possible that a larger ResNet architecture could have outperformed the DenseNet architectures that we tried. However, ResNet is inefficient in its parameters usage \cite{huang2016}, and thus a larger model was unable to fit on our GPUs.

\begin{table}[h]
  \caption{Effects of Neural Network Architectures}
  \label{tab:arch}
  \centering
  \begin{tabular}{lcc}
    \toprule
    Architecture & Validation Accuracy & Challenge Score\\
    \midrule
    VGG16 & 53.1\% & 472774.00\\
    ResNet50 & 67.3\% & 578552.21\\
    DenseNet121 & 75.3\% & 614221.42\\
    DenseNet161 & 78.7\% & 664645.25\\
  \bottomrule
\end{tabular}
\end{table}

\subsection{Class Imbalance and Optimization}
\begin{figure}
	\centering
	\includegraphics[width=3in, height=2.25in]{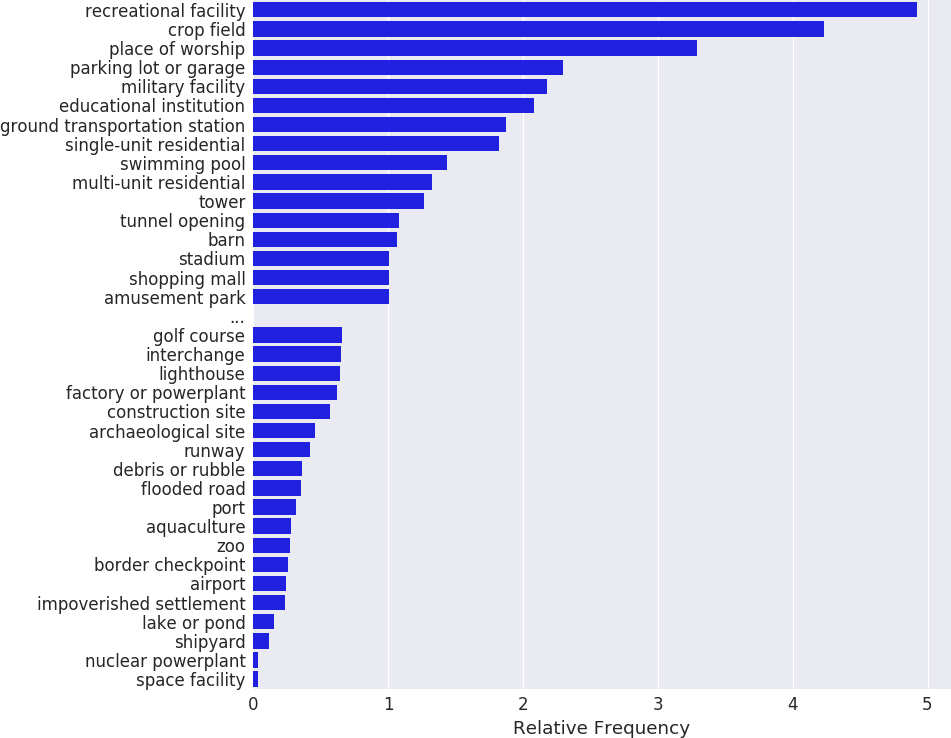}
	\caption{Relative frequencies of classes in dataset (with some omitted).}
	\label{fig:ClassRelativeFreq}
\end{figure}
Many classes were over-represented in the dataset, while others were severely under-represented (Figure \ref{fig:ClassRelativeFreq}), leading to a classic imbalanced class learning problem. There are many strategies to accommodate imbalanced classes \cite{Chawla2002, Drummond2003, Tang2009}.

Over-sampling and under-sampling yielded modest improvements in accuracy \cite{Chawla2009}. Skewed batches -- in which certain categories were completely missing for multiple batches -- were a larger problem. Since batch sizes were not always greater than the number of categories, we implemented a strategy in which categories were randomly sampled from the dataset with probability $$P = {P(y_j)}\log{\frac{1}{P(y_j)}}$$ where $y_j \in \{y_1, ... y_k\}$.
We also implemented a guarantee that a given category would be represented within $batch\_size / 63$ batches. This strategy did not have the gains in accuracy that we expected. However, it did stabilize our otherwise irregular training process, in which the network would randomly initialize in a state where the loss would keep increasing or would fail to leave a local minimum. This sampling process ensured that our model converged at a relatively smooth rate.

\section{Insights}

\subsection{Multiple-Instance Classification}
The FMoW challenge featured a unique problem. Each satellite image in the training data contained exactly one bounding box, but images in the test set contained multiple bounding boxes, each to be classified by our algorithm.

The naive approach involves simply classifying each bounding box in a testing image independently. A more exacting approach would take into consideration the relationship between bounding boxes in a testing image and their labels, since bounding boxes that are in the same image are necessarily geographically close on Earth. Such an approach would involve heuristic reasoning about what land usages are likely to be found near each other. For example, one could guess that it is unlikely for a golf course to be near an archaeological site. It would therefore be desirable to make a classifier that assigns low probability to the event that two bounding boxes in the same image would have the ``golf course" and ``archaeological site" labels. Since the training dataset  contains only one bounding box per image, assigning low probability to image combinations would be impossible to do purely through supervised learning. Possible solutions include generating surrogate secondary bounding boxes during training but asking the network to classify only the bounding box for which the true label is known, and creating a multiple-instance classifier equipped with a hand-engineered (not learned) matrix of expected class coincidences.

In the end, we decided to perform the naive, independent bounding-box classification due to time constraints.

Preliminary results from ongoing research demonstrates that more intelligent multiple-instance classification will increase the precision and recall across the categories of satellite imagery classification.

\section{Conclusion}
In this paper, we have identified limitations in an end-to-end deep learning task pipeline. Namely, we discuss limitations in obtaining, storing, processing, and modeling a large dataset of satellite imagery. We highlight the importance of maintaining a fast and efficient pipeline at all steps to ensure that tasks can be explored with quick iteration speeds and stability. Furthermore, we discuss limitations and possible solutions in relation to deep learning for image classification and present a few areas of future research.

\section*{Acknowledgment}
We gratefully acknowledge the support of Jonathan Chu, Nathan VanHoudnos, Scott McMillan, Oren Wright, and Hollen Barmer.

Copyright 2018 Carnegie Mellon University. All Rights Reserved.
This material is based upon work funded and supported by the Department of Defense under Contract No. FA8702-15-D-0002 with Carnegie Mellon University for the operation of the Software Engineering Institute, a federally funded research and development center.
References herein to any specific commercial product, process, or service by trade name, trade mark, manufacturer, or otherwise, does not necessarily constitute or imply its endorsement, recommendation, or favoring by Carnegie Mellon University or its Software Engineering Institute.
DM18-1297

\bibliographystyle{IEEEtran}
\bibliography{actual-references}

\end{document}